\documentclass[conference]{IEEEtran}
\IEEEoverridecommandlockouts
\usepackage{cite}
\usepackage{todonotes}
\usepackage{amsmath,amssymb,amsfonts}
\usepackage{algpseudocode}
\usepackage{algorithm}
\usepackage{graphicx}
\usepackage{textcomp}
\usepackage{subfig}

\def\BibTeX{{\rm B\kern-.05em{\sc i\kern-.025em b}\kern-.08em
    T\kern-.1667em\lower.7ex\hbox{E}\kern-.125emX}}

\newcommand{\ap}{$\mathcal{AP}$}
\newcommand{\budget}{$\mathcal{B}$}
\newcommand{\ds}{$\mathcal{DS}$}
\newcommand{\conv}{$conv$}
\newcommand{\sa}{\textit{Shadowing Agent}}

\begin{document}

\title{Shallow decision-making analysis in General Video Game Playing\\
%
}

\author{\IEEEauthorblockN{Ivan Bravi, Diego Perez-Liebana and Simon M. Lucas}
\IEEEauthorblockA{School of Electronic Engineering and Computer Science\\
Queen Mary University of London\\
London, United Kingdom \\
\textit{\{i.bravi, diego.perez, simon.lucas\}@qmul.ac.uk}
\and
\IEEEauthorblockN{
Jialin Liu}
\IEEEauthorblockA{Southern University of Science and Technology\\
Shenzhen, China\\
\textit{liujl@sustc.edu.cn}}
}
}

\maketitle

\begin{abstract}
The General Video Game AI competitions have been the testing ground for several techniques for game-playing, such as evolutionary computation techniques, tree search algorithms, hyper-heuristic-based or knowledge-based algorithms. 
So far the metrics used to evaluate the performance of agents have been win ratio, game score and length of games.
In this paper we provide a wider set of metrics and a comparison method for evaluating and comparing agents. The metrics and the comparison method give shallow introspection into the agent's decision-making process and they can be applied to any agent regardless of its algorithmic nature.
In this work, the metrics and the comparison method are used to measure the impact of the terms that compose a tree policy of an MCTS-based agent, comparing with several baseline agents.
The results clearly show how promising such general approach is and how it can be useful to understand the behaviour of an AI agent, in particular, how the comparison with baseline agents can help understanding the shape of the agent decision landscape.
The presented metrics and comparison method represent a step toward to more descriptive ways of logging and analysing agent's behaviours.
\end{abstract}


\section{Introduction}
General video game playing (GVGP) and General game playing (GGP) aim at designing AI agents that are able to play more than one (video) game successfully alone without human intervention. One of the early stage challenges is to define a common framework that allows the implementation and testing of such agents on multiples games. For this purpose, the General Video Game AI (GVGAI) framework~\cite{perez2018general} and General Game Playing framework~\cite{genesereth2005general,love2008general} have been developed. Competitions using the GVGAI and GGP frameworks have significantly promoted the development of a variety of AI methods for game-playing. Examples include tree search algorithms, evolutionary computation, hyper-heuristic, hybrid algorithms, and combinations of them. GVGP is more challenging due to the possibly stochastic nature of the games to be played and the short decision time. Five competition tracks have been designed based on the GVGAI framework for specific research purposes. The planning and learning tracks focus on designing an agent that is capable of playing several unknown games respectively with or without the forward model to simulate future game states. The level and rule generation tracks have the objective of designing AI programs that are capable of creating levels or rules based on a game specification.
Despite the fact that the initial purpose of developing GVGAI framework was to facilitate the research on GVGP, GVGAI and its game-playing agents have also been used in other application rather than just competitive GGP. For instance, the GVGAI level generation track has used the GVGAI game playing agents to evaluated the automatically generated game levels.
Relative algorithm performance \cite{nielsen2015general} has been used to understand how several agents perform in the same level. Although, no introspection into the agent behaviour or decision-making process was used so far.
 
The main purpose of this paper is to give a general set of metrics that can be gathered and logged during the agent's decision-making process to understand its in-game behaviour.
These are meant to be generic, shallow and flexible enough to be applied to any kind of agent regardless of its algorithmic nature.
Moreover we are also providing a generic methodology to analyse and compare game-playing agents in order to get an insight on how the decision-making process is carried out. This method will be later addressed as \textit{comparison method}.

Both the metrics and the comparison method will be useful in several applications.
It can be used for level generation: knowing the behaviour of an agent and what attracts it in the game-states space means that it can be used to measure how a specific level design suits a certain play-style therefore pushing the design to suit the agent in a recommender system fashion~\cite{machado2016shopping}. 
From a long term perspective, this can be helpful to understand a human player's behaviour and then personalise a level or a game to meet this player's taste or playing style.
Solving the dual problem is useful as well, in the process of looking for an agent that can play well a certain level design, disposing of reliable metrics to analyse the agent behaviour could significantly speed up the search.
Additionally, by analysing the collected metrics, it's possible to find out if a rule or an area of the game world is obsolete.
This can be also applied generally to the purpose of understanding game-playing algorithms, it's well known that there are black-box machine learning techniques that offer no introspection in their reasoning process, thus being able of comparing in a shallow manner, the decision-making process of different agents can help shed some light into their nature. A typical example is a neural network that given some input features outputs the action probability vector. With the proposed metrics and methodology it would be possible to make estimate its behaviour without actually looking at the agent playing the game and extracting behavioural information by hand.

The rest of this paper is structured as follows. In Section \ref{sec:back}, we provide a background on the GVGAI framework focusing in particular on the game-playing agents, three examples of how agent performance metrics have been used so far in scenarios other than pure game-play and an overview of MCTS-based agents. Then, we propose a comparison method, a set of metrics and an analysis procedure in Section \ref{sec:methods}. Experiments using these metrics are described in Section \ref{sec:ex_setup} and the results are discussed in Section \ref{sec:xp} to demonstrate how they provide a deeper understanding on the agent's behaviour and decision-making. Last, we draw final considerations and list possible future work in Section \ref{sec:conc}.

\section{Background}\label{sec:back}

  \subsection{General Video Game AI framework}
 The General Video Game AI (GVGAI) framework~\cite{perez2018general} has been used for organising GVGP competitions at several international conferences on games or evolutionary computation, for research and education in worldwide institutions. The main GVGAI framework is implemented using \emph{Java} and \emph{Python}. A Python-style Video Game Description Language (VGDL)~\cite{ebner2013towards,schaul2013video} is developed to make it possible to create and add new games to the framework easily. The framework enables several tracks with different research purposes. The objective of the single-player~\cite{perez20162014} and two-player planning~\cite{gaina20172016} tracks is to design an AI agent that is able to play several different video games respectively alone or with another agent. With access to the current game state and the forward model of the game, a planning agent is required to return a legal action in a limited time. Thus, it can simulate games to evaluate an action or a sequence of actions and get the possible future game state(s). However, in the learning track, no forward model is given, a learning agent needs to learn in an trial-and-error way. There are two other tracks based on the GVGAI framework which focus more on game design: the rule generation~\cite{khalifa2016general} and the level generation~\cite{khalifa2017rulegen}. In the rule generation track, a competition entry (generator) is required to generate game rules (interactions and game termination conditions) given a game level as input, while in the level generation track, an entry is asked to generate a level for a certain game. The rule generator or level generator should be able to generate rules or levels for any game given a specified search space.

\subsection{Monte Carlo Tree Search-based agents}
Monte-Carlo Tree Search (MCTS) has been the state-of-the-algorithm in game playing~\cite{browne2012survey}. 
The goal of MCTS is to approximate the value of the actions/moves that may be taken from the current game state. 
MCTS builds iteratively a search tree using Monte Carlo sampling in the decision space and the selection of the node (action) to expand is based on the outcome of previous samplings and on a Tree Policy. A classic Tree Policy is the Upper Confidence Bound (UCB)~\cite{auer2002finite}. The UCB is one of the classic multi-armed bandit algorithms which aims at balancing between exploiting the best-so-far arm and exploring more the least pulled arms. Each arm has an unknown reward distribution. In the game-playing case, each arm models a legal action from the game state (thus a node in the tree), a reward can be the game score, a win or lose of a game, or a designed heuristic. The UCB Tree Policy selects to play the action (node) $a^*$ such that $
a^*=\arg \max_{a \in A}\ \bar x_{a} + \sqrt{\frac{\alpha \ln{n}}{n_a}}$, where $A$ denotes the set of legal actions at the game state, $n$ and $n_a$ refers to the total number of plays and the number of times that the action $a$ has been played (visited), $\alpha$ is called exploration factor. 

The GVGAI framework provides several sample controllers for each of the tracks. For instance, the \emph{sampleMCTS} is a vanilla implementation of MCTS for single-player games, but performs finely on most of the games. M. Nelson~\cite{nelson2016investigating} tests the \emph{sampleMCTS} on more than sixty GVGAI games, using different amounts of time budget for planning at every game tick, and observes that this implementation of MCTS is able to reduce the loss rate given longer planning time.
More advanced variants of MCTS have been designed for playing a particular game (e.g., the game of Go~\cite{silver2016mastering,silver2017mastering}), for general video game playing (e.g., \cite{perez20162014,soemers2016enhancements}) or general game playing (e.g., \cite{mehat2010combining}). Recently, Bravi el al.~\cite{bravi2017master} custom various heuristics particularly for some GVGAI games, and Sironi el al.~\cite{sironi2018self} design several Self-Adaptive MCTS variants which use hyper-parameter optimisation methods to tune on-line the exploration factor and maximal roll-out depth during the game playing.

  \subsection{Agent performance evaluation}
  Evaluating the performance of an agent is sometimes a very complex task depending on how the concept of performance is defined. 
In the GVGAI planning and learning competitions, an agent is evaluated based on the the amount of games it wins over a fixed number of trials, the average score that it gets and the average duration of the games.
Sironi et al. \cite{sironi2018self} evaluate the quality of their designed agents using a heuristic which combines the score obtained eventually giving an extra bonus or penalty depending on whether the agent could reach a winning state or a losing state, respectively.
The GVGAI framework has also been used for purposes other than the ones laid out by the competition tracks.
Bontrager et al. \cite{bontrager2016matching} cluster some GVGAI single-player and two-player games using game features and agent performance extracted using the playing data by the single-player and two-player planning competition entries, respectively.
In particular, the performance of an agent, represented by win ratio in \cite{bontrager2016matching}, is used to cluster the games in four groups: games easy to win, hard games, games that MCTS agent can play well and games that can be won by a specific set of agents.
The idea behind that work is interesting although the clustering results in three small sized groups and a very large one. This suggests that using more introspective metrics could help clustering the games more finely.

GVGAI has also been used as test bed for evolving MCTS tree policies (in the form of a mathematical formula for decision making) for specific games~\cite{bravi2017master}. 
\cite{bravi2017master} consists in evolving Tree Policies (formulae) using Genetic Programming, the fitness evaluation is based on the performance of an MCTS agent which uses the specific tree policy.
Once again, the informations logged and used from the playthrough by the fitness function were a combination of win ratio, average score and average game-play time, in terms of the number of game ticks. Unfortunately no measurement was made about the robustness of the agent's decision-making process of which could have been embedded in the fitness function to possibly enhance the evolutionary process.
In the recent Dagstuhl seminar on AI-Driven Game Design, game researchers have envisioned a set of features to be logged during game-play, divided into four main groups: direct logging features, general indirect features, agent-based features and interpreted features~\cite{measures2018dagstuhl}.
A preliminary example of how such features can be extracted and logged in the GVGAI framework has also been provided~\cite{measures2018dagstuhl}. Among the direct logging features, we can find some kind of game information that don't need any sort of interpretation, few examples are: game duration, actions log, game outcome and score. 
Instead, these features are listed in the general indirect features which require some degree of interpretation or analysis of the game state such as the entropy of the actions, the game world and the game state space.
The agent-based features gather information about the agent(s) taking part to the game, for example about the agent surroundings, the exploration of the game-state space or the convention between different agents. Finally, the interpreted features are based on metrics already defined in previous works such as drama and outcome uncertainty~\cite{browne2010evolutionary} or skill depth~\cite{liu2017evolving}.

\section{Methods}\label{sec:methods}
  
  This section first introduces a set of metrics that can potentially be extracted from any kind of agent regardless of its algorithmic nature, aiming at giving an introspection of the decision-making process of a game-playing agent in a shallow and general manner (Section \ref{metrics}). Then we present a method to compare the decisions of two distinct game-playing agents under identical conditions using the metrics introduced previously. As described in \cite{holmgaard2016evolving} the decision-making comparison can be done at growing levels of abstraction: action, tactic or strategic level. Our proposed method compares the decision-making at the action level. Later, we design a scenario in which the metrics and the comparison method are used to analyse the behaviour of instances of an MCTS-agent using different tree policies comparing them to agents with other algorithmic natures. Finally we describe the agents used in the experiments.
  
In this paper, the following notations are used.
A \textit{playthrough} refers to a complete play of a game from beginning to end. The set of available actions is denoted as $\mathcal{A}$ being $N=|\mathcal{A}|$, $a_i$ refers to the $i^{th}$ action in $\mathcal{A}$. A \textit{budget} or \textit{simulation budget} is either the amount of forward-model calls the agent can make at every game tick to decide the next action to play or the CPU-time that the agent can take. The fixed budget is later addressed as \budget.
  
    \subsection{Metrics} \label{metrics}
	The metrics presented in this paper are based on two simple and fairly generic assumptions: (1) for each game tick the agent considers each available action $a_i$ for $n_i$ times; (2) for each game tick the agent assigns a value $v(a_i)$ to each available action. In this scenario the agents are designed to operate on a fixed budget \budget~ in terms of real time or number of forward model calls, which allows for a fair comparison making the measurements comparable between each other.

Due to the stochastic nature of an agent or a game, it is sometimes necessary to make multiple playthroughs for evaluation.
The game id, level id, outcome (specifically, win/loss, score, total game ticks) and available actions at every game tick are logged for each playthrough.
Additionally, for each game tick in the playthrough, the agent is going to provide the following set of metrics:
      \begin{itemize}
      \item $a^*$: the recommended action to be played next;
      \item $\overline{p}$: probability vector where $p_i$ represents the probability of considering $a_i$ during the decision-making process;
      \item $\overline{v}$: vector of values $v_i \in \mathcal{R}$ where $v_i$ is the value of playing $a_i$ from the current game state, $v^*$ is the highest value which implies it being associated with $a^*$. Whenever the agent doesn't actually have such information about the quality of $a_i$ then $v_i$ should be NaN;
      \item $b$: represents the ratio of the budget consumed over the fixed available budget \budget, $b \in [0,1]$ where $0$ and $1$ respectively mean that either no budget or the whole \budget~ was used by the agent;
      \item \conv: convergence, as the budget is being used is likely for the current $a^*$ to fluctuate, \conv~ is the ratio of budget used over \budget~ when $a^*$ is stable. It means that any budget used after \conv~ hasn't changed the recommended action. \conv~$ \in [0,b]$.
      \end{itemize}

It is notable that most of the agents developed for the GVGAI try to consume as much budget as possible, however this is not necessarily a good trait of the agent, being able to log the amount of budget used and distinguish between a budget-saver and a budget-waster can give an interesting insight on the decision-making process especially on the confidence of the agent. Since this set of metrics tries to be as generic as possible, we shouldn't limit the metrics because of the current agent implementations.
The vectors $\overline{p}$ and $\overline{v}$ can be inspected to portray the agent preference over $\mathcal{A}$. The vector $\overline{p}$ can also be used during the debug phase of designing an agent to see whether it actually ever considers all the available action.


Generally different agents reward actions differently, therefore it is not possible to make a priori assumptions on the range or the distribution over values. Although the values in $\overline{v}$ allow at the very least to rank the actions and moreover to get informations about their boundaries and distributions (guaranteed a reasonable amount of data) a posteriori. Furthermore, it is possible to follow the oscillation of such values through the game-play highlighting critical portions of it. For example, when the $v_i$ are similar (not very far apart from each other considering the value bounds logged) and generally high then we can argue that the agent evaluates all actions as good ones. On the contrary if the values are generally low, the agent is probably struggling in a bad game scenario.


    \subsection{Comparison method}\label{sec:comparison}
    
    Comparing the decisions made by different agents is not a trivial matter especially when their algorithmic nature can be very different. The optimal set-up under which we can compare their behaviour is when they are provided the same problem or scenario under exactly same conditions. This is sometimes called \emph{pairing}. 
    We propose the following experimental set-up: a meta-agent, called \sa, instantiates two agents: the \textit{main} agent, and the \textit{shadow} agent.
    For each game tick the \sa~behaves as a proxy and feeds the current game state to each of the agents which will provide the next action to perform as if it was a normal GVGAI game-play execution. Both these agents have a limited budget.
    Once both main and shadow agent behaviours are simulated, the \sa~takes care of logging the metrics described previously from both agents and then returns to the framework the action chosen by the main agent. In this way the actual avatar behaviour in the game simulated is consistent with the main agent and the final outcome represents its performance. In the next sections we are going to use the superscripts $^m$ and $^s$ for a metric respectively relative to the main agent or the shadow agent.
    A typical scenario would be comparing how very radically different agents such as: a Random agent, a Monte-Carlo Search agent, a One-Step Look Ahead agent and an MCTS-based agent. Under this scenario, comparing each single coupling of agents will result in producing a matrix of comparisons. All the informations on how the agents extract the metrics described previously are detailed in Section \ref{agents}.

    \subsection{Analysis Method}
    
\begin{figure}[htbp]
  \centering
    \includegraphics[width=0.8\linewidth]{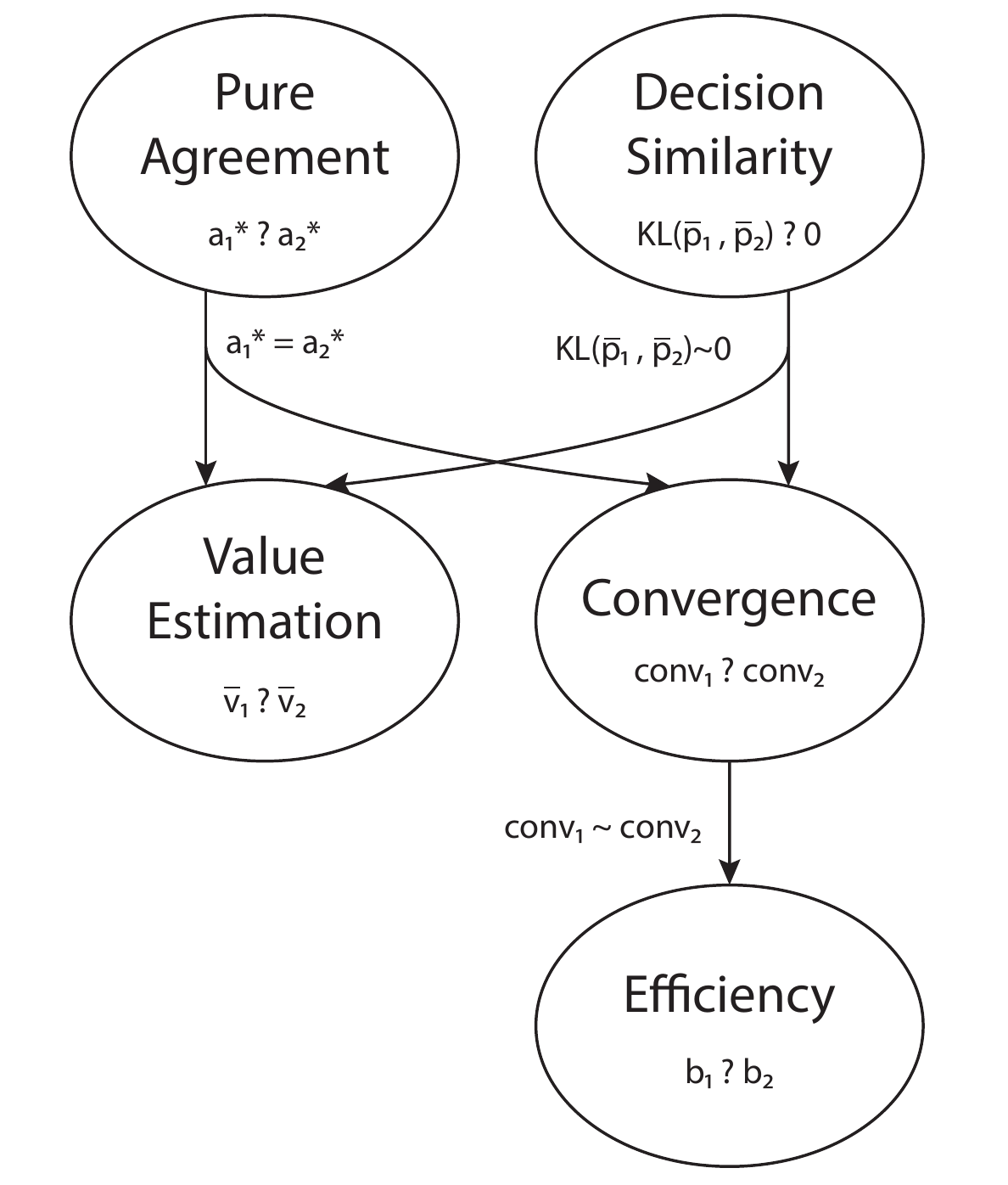}
    \caption{\label{analysis-graph}The decision graph to compare agents' behaviours.}
\end{figure}

	We are going to analyse these agents' behaviours in few games, for each game we are going to run all the possible couplings of main agent and shadow agent, for each couple we are going to run $N_p$ playthroughs and, finally, for each playthrough we are going to save the current metrics for both main and shadow agents. It's worth remembering that each playthrough has its own length, thus playthrough $i$ will have length $l_i$.
	This means that in order to analyse and compare behaviours we need a well structured methodology to slice data appropriately. Our proposed method is represented in Figure \ref{analysis-graph}. The first level of comparison is done at the action level, we can measure two things: \textit{Agreement Percentage} \ap, percentage of times the agents agreed on the best action averaged across the several playthroughs;
	and \textit{Decision Similarity} \ds, the average symmetric Kullback-Leibler divergence of the two probability vectors $\overline{p}^m$ and $\overline{p}^s$. When \ap~ is close to $100\%$ or \ds~$\sim0$ we have two agents with similar behaviours, at this point we can step to the next level of comparison: \textit{Convergence}, we compare \conv$^m$ and \conv$^s$ to see if there is a faster converging agent; and \textit{Value Estimation}, this level of comparison is thorny, in fact each agent has its own function for evaluating a possible action, for this step we recommend using these values to rank the actions using them as preference evaluation. \textit{Convergence} can highlight both the ambiguity of the surrounding game states or the inability of the agent to recognise important features. If the agents have a similar \conv~ values we can then take a look at the \textit{Efficiency}. This value represents the average amount of budget used by the agent.
    
    To summarise, once two agents with similar \ap~ or \ds~ are found, the next comparison levels highlight the potential preference toward the fastest converging and most budget-saver one.
	
    \section{Experimental set-up} \label{sec:ex_setup}
    In this section, we show how a typical experiment could be run using the metrics and methods introduced previously. Each experiment is run over the following games in order to have diverse scenarios that can highlight different behaviours:
  \begin{itemize}
  \item Aliens: a game loosely modelled on the Atari 2600's Space Invaders, the agent on the bottom of the screen has to shoot the incoming alien spaceships from above avoiding their blasts;
  \item Brainman: the objective of the game is for the player to reach the exit, the player can collect diamonds to get points and push keys into doors to open them;
  \item Camel Race: the player, controlling a camel, has to reach the finish line before the other camels whose behaviour is part of the design of the game;
  \item Racebet: in the game there are few camels racing toward the finish line, each has a unique colour, in order to win the game the agent has to position the avatar on the camel with a specific colour;
  \item Zenpuzzle: the level has two different types of floor tiles, one that can be always stepped on and a special type that can be stepped on no more than once. The agent has to step on all the special tiles in order to win the game.
  \end{itemize}
  Further details on the games and the framework can be found at www.gvgai.net .
  The budget given to the agents is a certain number of forward-model calls which is different than the real time constraints used in the GVGAI competitions.
  We made this decision in order to get more robust data across different games, in fact the number of forward model calls that can be executed in the 40 ms can drastically vary changing the game, sometimes from hundreds to thousands.
  
  This experiment consists in running the comparisons between the MCTS-based agents that use all possible prunings $h' \in \mathcal{H}$ as tree policy generated from $h$ (cf. \eqref{A0}, variables summarised in Table \ref{variables}), and the following agents: Random, One-Step Look Ahead, and Monte-Carlo Search.
  \begin{equation}
    \label{A0}
    h = min(D_{MOV}) \cdot min(D_{NPC}) + \frac{\lvert max(R)\rvert}{\sum D_{NPC}}
  \end{equation}
   \begin{table}[h]
    \centering
    \caption{Variables used in the heuristic (cf. \eqref{A0}).}
    \renewcommand{\arraystretch}{1.4}
    \setlength\tabcolsep{1.4pt}
    \begin{tabular}{c|l}
    \hline
    \textbf{Notation} & \multicolumn{1}{c}{\textbf{Description}} \\
    \hline
    $max(R)$ & Highest reward among the simulations that visit current node \\
    $min(D_{MOV})$ & Minimum distance from a movable sprite \\
    $min(D_{NPC})$ & Minimum distance from an NPC \\
    $sum(D_{NPC})$ & Sum of all the distances from NPCs \\
    \hline
    \end{tabular}
    \label{variables}
  \end{table}
  In this work, each pair of agents is tested over 20 playthroughs of the first level of each game, all the agents were given a budget of $700$ forward-model calls. The budget was decided looking at the average number of forward-model calls done in all the GVGAI games by the Genetic Programming MCTS (GPMCTS) agent with a time budget of 40 ms, same as in the competitions. The GPMCTS agent is an MCTS agent with customisable Tree Policy as described in \cite{bravi2017master}.
    
    \subsection{Comparison method for MCTS-based agents}
    MCTS-based agents can be tuned and enhanced in many different ways, a wide set of hyper-parameters can be configured differently, one of the most crucial components is the tree policy. The method we propose gradually prunes the tree policy heuristic in order to isolate bits of \eqref{A0}. 
    Evaluating the similarity of two tree policies is a rather complex task, it can be roughly done by analysing the difference between their values given a point in their search domain. This approach is not optimal, supposing we want to analyse two functions $f$ and $g$ where $g = f + 10$, their values will never be the same but when applied to the MCTS scenario they would perform exactly the same. Actually, what matters is not the exact value of the function but the way that two points in the domain are ordered according to their evaluations. In short, being $\mathcal{D}$ the domain of the functions $f$ and $g$ and $p_1,p_2 \in \mathcal{D}$ what matters is that both the following conditions $f(p_1) \ge f(p_2)$ and $g(p_1) \ge g(p_2)$ hold true.
The objective is to understand how each term in \eqref{A0} used in the tree policy of an MCTS agent impacts the behaviour of the whole agent. Given $h$, thus \eqref{A0} used as tree policy, let $\mathcal{H}$ be the set of all possible prunings (therefore functions) of the expression tree associated to $h$. This method applies the metrics and the comparison method introduced previously and it consists in running all possible couples $(A_m,A_s) \in \mathcal{AG}\times\mathcal{AG}$ where the agent $A_m$ is the main agent and $A_s$ is the shadow agent, the set $\mathcal{AG}$ contains one instance of MCTS-based agent for each tree policy in $\mathcal{H}$ and the following agents: Random, One-Step Look Ahead, Monte-Carlo Search.
    In this way it is possible to get a meaningful evaluation of how different equations might result in suggesting the same action, or not, for all the possible comparisons of the equations in $\mathcal{H}$ but also how they compare to the other reference agents.

      \subsection{Agents}
      \label{agents}
      In this section, we give the specifications of the agents used and the way they link each metric to their algorithmic implementation. These agents are going to be used in the experiments and they can be used as examples of how algorithmic informations can be interpreted and manipulated to get the metrics described previously. Most agents use \textit{SimpleStateHeuristic} which evaluates a game state according to the win/lose state, the distance from portals and the number of NPCs. It rewards best winning states with no NPCs and where the position of the player is closest to a portal.
      None of the agents was chosen for its performance, the point of using these agents is that theoretically they can represent very different play styles: completely stochastic, very short-sighted, randomly long-sighted, generally short-sighted.
      	\subsubsection{Random}
	The random agent has a very straightforward implementation: given the set of available actions, it picks an action uniformly at random.
	\begin{itemize}
      \item $\overline{p}$: since the action is picked uniformly $p_i = 1/|\mathcal{A}|$;
      \item $\overline{v}$: each $v_i$ is set to NaN;
      \item $b=0$, since no budget is consumed to return a random action;
      \item \conv~is always $0$ for the same reason of $b$.
      \end{itemize}
      
      \subsubsection{One-Step Look Ahead}
    The agent makes a simulation for each of the possible actions, and evaluates the resulted game state using the \textit{SimpleStateHeuristic} defined by the GVGAI framework. The action with the highest values is going to be picked as $a^*$.
	\begin{itemize}
      \item $\overline{p}$: $p_i = 1/|\mathcal{A}|$ since each action is picked once;
      \item $\overline{v}$: each $v_i$ corresponds to the evaluation given by the \textit{SimpleStateHeuristic} initialized with current game state and compared to the game state reached via action $a_i$;
      \item $b$ is always $\frac{|\mathcal{A}|}{sb} $;
      \item \conv~ varies and corresponds to the budget ratio when the best action is simulated.
      \end{itemize}
      
      \subsubsection{Monte-Carlo Search}
	The Monte-Carlo Search agent performs a Monte-Carlo sampling of the action-sequence space following 2 constraints: the sequence is not longer than 10 and only the last action can bring to a termination state.
	\begin{itemize}
      \item $\overline{p}$: considering $n_i$ as the number of times action $a_i$ was picked as first action and $N = \sum_{i=0}^{|\mathcal{A}|}n_i$ then $p_i = \frac{n_i}{N}$;
      \item $\overline{v}$: each $v_i$ is the average evaluation by the $\textit{SimpleStateHeuristic}$ initialized with the current game state compared to each last game state reached by every action sequence started from $a_i$;
      \item $b$ is always 1, since the agent keeps simulating until the end of the budget;
      \item \conv~corresponds to the ratio of budget used at the moment the action with the highest $v_i$ last changed.
      \end{itemize}
      
      \subsubsection{MCTS-based}
	The MCTS-based is an implementation of MCTS with uniformly random roll-outs to a maximum depth of 10.
    The tree policy used can be specified when the agent is initialised, therefore the reader should not suppose UCB1 as the tree policy, whereas the heuristic used to evaluate game states is a combination of the score plus an eventual bonus/penalty for a win/lose state.
	\begin{itemize}
      \item $\overline{p}$: considering $n_i$ as the number of visits for $a_i$ at the root node of the search tree and $N$ as the number of visits at the root node then $p_i = \frac{n_i}{N}$;
      \item $\overline{v}$: each $v_i$ is the heuristic value associated to $a_i$ at the root node;
      \item $b=1$, since the agent keeps simulating until the budget is used up;
      \item \conv~corresponds to the ratio of budget used when the action with the highest $v_i$ last changed in the root node.
      \end{itemize}
  
\section{Experiments}\label{sec:xp}
    
\begin{figure*}
    \centering
    \subfloat[Aliens Pure Agreements]{
  	  \label{bb:m_aliens}
      \centering
      \includegraphics[width=.43\linewidth]{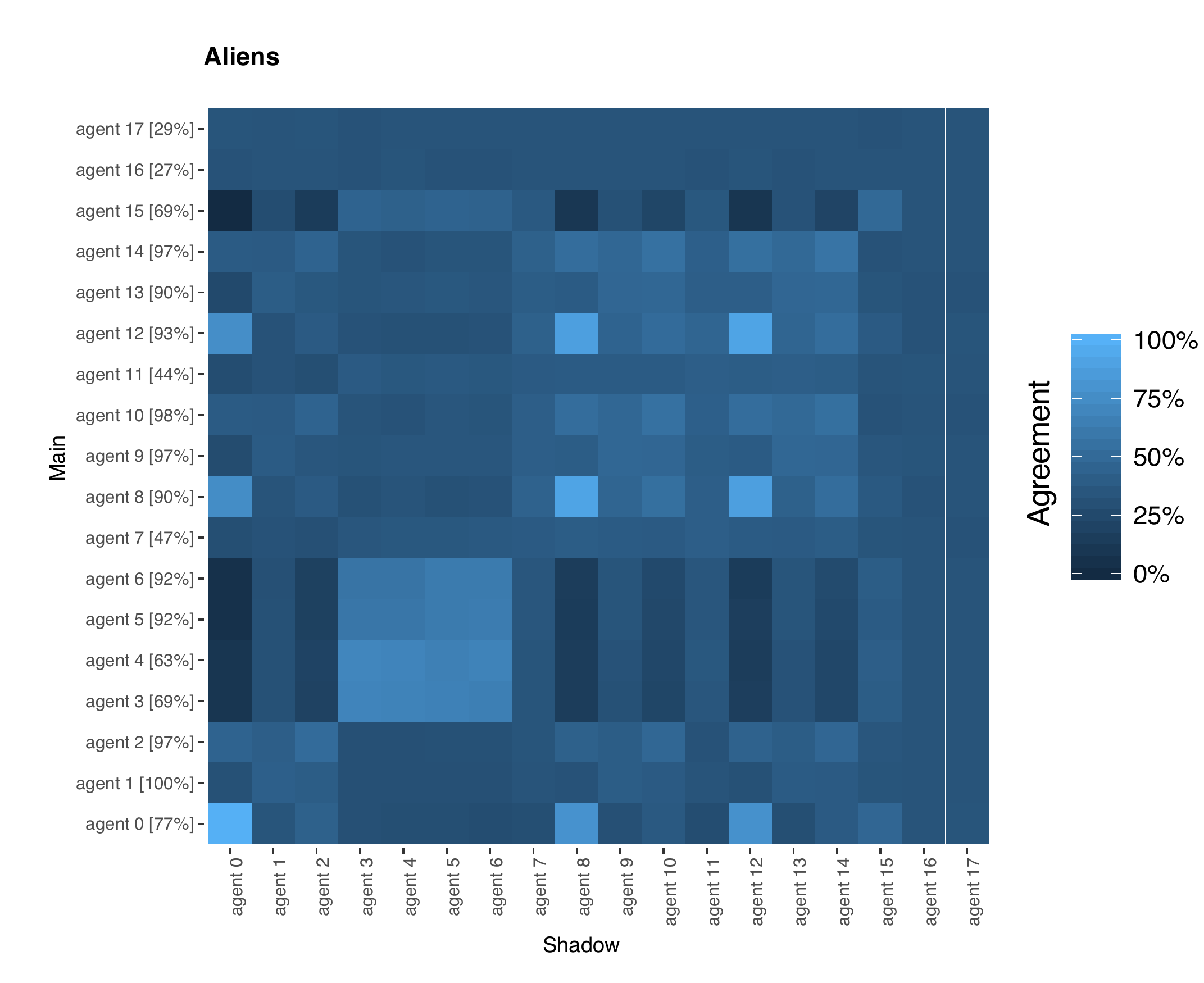}
    }
    \subfloat[Zenpuzzle Decision Similarities]{
      \label{bb:zen_kl}
      \centering
      \includegraphics[width=.43\linewidth]{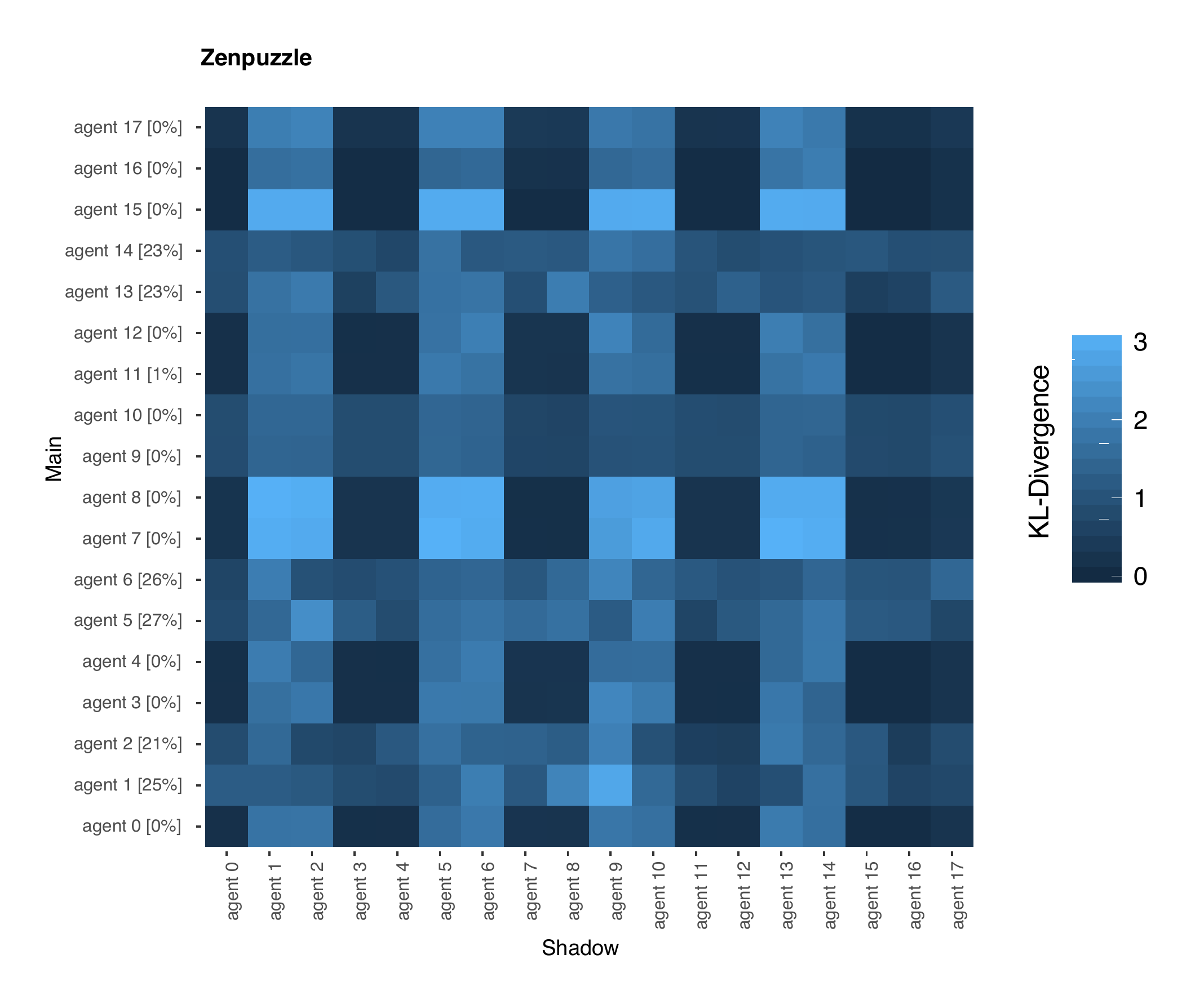}
    }
    \caption{\label{bigboy} Results of two comparison scenarios between all the agents in Table \ref{agents_table}. In Figure \ref{bb:m_aliens} we have the comparison using the \textit{Pure Agreement} method, the values from dark blue to light blue represent the agreement percentage (the lighter the higher). Instead in Figure \ref{bb:zen_kl} light blue represents very diverging action probability vectors while the darkest blue is for the case those are identical. The vertical and the horizontal dimensions of the matrix represent the main and shadow agent, respectively, in the comparison process. The main agent's win percentage is specified between square brackets in its label on the vertical axis.}
  \end{figure*}

    \begin{table}[h]
    \centering
    \caption{Agents used in experiments and their ids.}
    \label{agents_table}
    \renewcommand{\arraystretch}{1.4}
    \scriptsize
    \begin{tabular}{|c|l|}
      \hline
        Id & Agent\\
      \hline
        0 & MCTS + $\frac{1}{\sum D_{NPC}}$ \\
        1 & MCTS + $\lvert max(R)\rvert$ \\
        2 & MCTS + $\frac{\lvert max(R)\rvert}{\sum D_{NPC}}$ \\
        3 & MCTS + $min(D_{NPC})$ \\
        4 & MCTS + $min(D_{NPC}) + \frac{1}{\sum D_{NPC}}$ \\
        5 & MCTS + $min(D_{NPC}) + \lvert max(R)\rvert$ \\
        6 & MCTS + $min(D_{NPC}) + \frac{\lvert max(R)\rvert}{\sum D_{NPC}}$ \\
        7 & MCTS + $min(D_{MOV})$ \\
        8 & MCTS + $min(D_{MOV}) + \frac{1}{\sum D_{NPC}}$ \\
        9 & MCTS + $min(D_{MOV}) + \lvert max(R)\rvert$ \\
        10 & MCTS + $min(D_{MOV}) + \frac{\lvert max(R)\rvert}{\sum D_{NPC}}$ \\
        11 & MCTS + $min(D_{MOV}) \cdot min(D_{NPC})$ \\
        12 & MCTS + $min(D_{MOV}) \cdot min(D_{NPC}) + \frac{1}{\sum D_{NPC}}$ \\
        13 & MCTS + $min(D_{MOV}) \cdot min(D_{NPC}) + \lvert max(R)\rvert$ \\
        14 & MCTS + $min(D_{MOV}) \cdot min(D_{NPC}) + \frac{\lvert max(R)\rvert}{\sum D_{NPC}}$ \\
        15 & One-Step Look Ahead \\
        16 & Random \\
        17 & Monte-Carlo Search\\
      \hline
    \end{tabular}
  \end{table}
    Table \ref{agents_table} summarises the agents used in the experiments and the ids assigned to them. Multiples MCTS agents using different tree policies have been tested.
Figure \ref{bigboy} illustrates an example of agreement percentage \ap~and another of decision similarity \ds~between the main agent and the shadow agent on two tested games. An important fact to remember when looking at Figure \ref{bb:m_aliens} is that the probability of two random agents agreeing on the same action is $\frac{1}{|\mathcal{A}|}$. Therefore, when looking at the \ap~we should take into account and analyse what deviates from $\frac{1}{|\mathcal{A}|}$. The game Aliens is the only game where the agent has three available actions, the rest of the game is played with four available actions.
The bottom-right to top-left diagonal in the matrix represents the \ap~that the agent has with itself, this particular comparison has a intrinsic meaning: it shows the coherence of the decision-making process, the higher the agreement the more consistent is the agent. This feature can be highlighted even more clearly looking at the \ds~ where the complete action probability vectors are compared.

This isn't necessarily always good feature especially in competitive scenarios where a mixed strategy could be advantageous, but it's a measure of how the search process is consistent with its final decision. Picturing the action-sequence fitness landscape, a high \ap~implies that the agent shapes it in a very precise and sharp definition being able to identify consistently a path through it. In the scenarios where a lot of navigation of the level is necessary, there might be several way to reach the same end goal, this will result in the agent having a lower self-agreement.

The KL-Divergence measure adopted for \ds~hilights how distinct are the decision making processes of each agent.  Using this approach we would then expect much stronger agreement along the leading diagonals of all the comparison matrices as Figure \ref{bb:zen_kl}.  Conversely, we would also expect a much clearer distinction between agents with genuinely distinct policies.

  \textbf{Aliens.}
  The game Aliens is generally easy to play, the Random agent can achieve a win rate of 27\%, and the MCTS alternatives achieve win rates varied from 44\% to 100\%. So there are clearly some terms of the equation used in tree policy which matter more than others. The best performing agent is the agent $0$ with a perfect win rate, which uses a very basic policy and chooses the action that maximises the highest value found, it's a greedy agent.
An interesting pattern is observed in Figure \ref{bb:m_aliens}: the agents $0$, $8$ and $12$ all share the same term $\frac{1}{\sum D_{NPC}}$ alone or together with $min(D_{MOV})$ it gives stability to the decisions taken. This is even clearer looking at the \ds~ value which are respectively 0, 0.067 and 0.07 . Agent $12$, the one with the best combination of \ap and win rate, is driven by a rather peculiar policy: the first term maximises the combined minimal distance from NPCs (aliens) and movable objects (bullets), the second term minimises the sum of the distances from NPCs. This translates into a very clear and neat game-playing strategy: stay away from bullets and kill the aliens (being the fastest way to reduce $\sum D_{NPC}$). This agent is not only very strong with a 93\% win rate, but also extremely fast in finding its preferred action with an average \conv$=0.26$. 
Even the win rate of agent $15$ is not one of the best ones, the $b$ metric highlights how an agent as $11$ is intrinsically flawed. In fact, even if agent $11$ constantly consumes all the budget at its disposal ($b=1$) it gets a win rate of just 44\% whilst agent $15$ with a $b<0.006$ is able to get a 69\% win rate.

  \textbf{Brainman.}
  This game is usually very hard for the AIs, the best one from the batch has a win rate of 31\%. Looking at the data we have noticed a high concentration of \ap~around 50\% for all combination of agents from 7 to 10, this is even clearer looking at the \ds data which is consistently below 0.2. When the policy contains the term $min(D_{MOV})$ not involved in any multiplication the agent is more consistent in moving far away from moving objects. Unfortunately that is exactly a behaviour that will never allow the agent to win, in fact, the key to open the door with the goal is the only movable object in the game.
      
  \textbf{Camelrace.}
  The best way to play Camelrace is easy to understand: keep moving right until reaching the finish line. Looking into the comparison matrix \ap~for this game, we've noticed how there's a big portion of it (agents from $3$ to $14$) where the agents consistently agree most of the time (most values over 80\%). What is interesting to highlight is how only that clustering with an \ap$=100$ (agents $8$ and $7$) can hit a win rate of 100\% which is further highlighted by \ds that is $0$. This is due to the fact that even just few wrong actions can backfire dramatically. In fact in the game there's an NPC going straight right thus wasting few actions means risking to be overcome by it and lose the race, therefore coherence is extremely important. 
  
 \begin{figure}[htbp]
  \centering
    \includegraphics[width=0.9\linewidth]{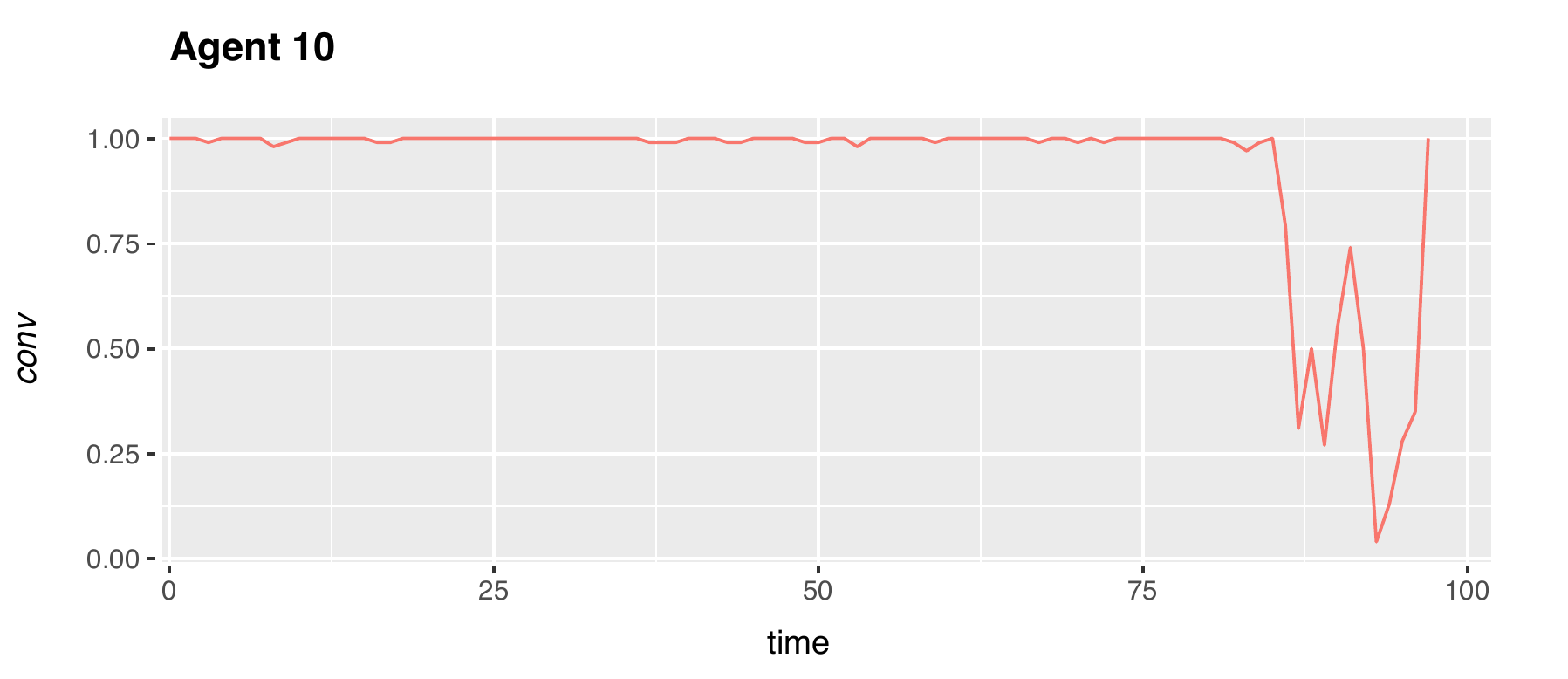}
    \caption{\label{racebet2_dp}The average $conv$ in the game Racebet2 for the agent $10$ throughout the plays. It shows how the agent doesn't clearly have a preference over the actions until the end of the game when the value drastically drops.}
\end{figure}
  \textbf{Racebet2.}
  The \ap~values for this game are harder to read, the avatar can move only in a very restricted cross-shaped area and its interaction with the game elements is completely useless until the end of the playthrough when the result of the race is obvious to the agent. This is clearly expressed by the average convergence value during the play for agent $10$ shown in Figure \ref{racebet2_dp}. Agent $10$ can not make up his mind consuming all the budget before settling for $a^*$ (\conv~$=1$), it keeps happening until the very end of the game when it has a drastic drop of \conv meaning that the agent is now able to swiftly decide the preferred action. Potentially, an agent could stand still for most of the game and move just during the last few frames of the game.
  This overall irrelevance of most actions during the game is exemplified by an almost completely flat value of \ap~ for most agent couples around $25\%$.

  
  \textbf{Zenpuzzle.}
  This is a pure puzzle game where to win the game is not sufficient following the rewards. The \ap~ values are completely flat, in this case the pure agreement doesn't provide any valuable information. However, as we can see in Figure \ref{bb:zen_kl}, the KL-divergence is more expressive to catch decision making differences and we can notice that generally being less consistent with itself can eventually take to perform the crucial right action to fill the whole puzzle. This is a perfect scenario to show a limit of \ap, there are several agents to win a game every four but without comparing the full action probability vector we couldn't have highlighted this crucial detail.

\section{Conclusion and Future Work}\label{sec:conc}
We have presented a set of metrics that can be used to log the decision-making process of a game-playing agent using the General Video Game AI framework. Together with these metrics, we also introduced a methodology to compare agents under the same exact conditions, both are applicable to any agent regardless of their actual implementation and the game they are meant to play.
The experimental results have demonstrated how combining such methods and metrics make it possible to have a better understanding on the decision-making process of the agents. In several occasions we have seen how the measuring the agreement between a simple and not necessarily well-performing agent and the target agent, can shed some light on the implicit intentions of the latter. Such approach holds the potential for developing a set of agents with a specific well-known behaviour that can be used to analyse, using the comparison method introduced, another agent's playthrough. They could be used as an array of shadow agents, instead of a single one, and measure during the same play if and how much the behaviour of the main agent resembles that of the shadow agents. 
Progressively pruning the original Tree Policy we have seen how it was possible to decompose it in simple characteristic behaviours with extremely compact formulae: fleeing a type of objects, maximising the score, killing NPCs. Recognising them has been proven helpful to then understand the behaviour of more complex formulae whose behaviour is not possible to be expected a-priori.

Measuring the \conv~has shown how it is possible to go beyond the sometimes-too-sterile win rate and to use both metrics to distinguish between more and less efficient agents.
The game Zenpuzzle has clearly shown that the current set of metrics is not sufficient.
The implementation of the \sa~and the single agents compatible with it will be released as open source code after the publication of this paper, together with the full set of comparison matrices, at www.github.com/ivanbravi/ShadowingAgentForGVGAI .
In future work the metrics can be extended to represent additional information about the game states explored by the agent, such as the average events triggered, average counter for each game element just to name few as examples, but also more features from the sets envisioned in \cite{measures2018dagstuhl}.


\bibliographystyle{IEEEtran}
\bibliography{bibliography}

\end{document}